\def\year{2019}\relax
\documentclass[letterpaper]{article} 
\usepackage{aaai19}  
\usepackage{times}  
\usepackage{helvet} 
\usepackage{courier}  
\usepackage[hyphens]{url}  
\usepackage{graphicx} 
\usepackage{graphicx}  
\usepackage{latexsym}
\usepackage{amsmath}
\usepackage{amssymb}
\usepackage{multirow}
\usepackage{algorithm}
\usepackage{algorithmic}
\usepackage[utf8]{inputenc} 
\usepackage{url}            
\usepackage{booktabs}       
\usepackage{amsfonts}       
\usepackage{array}
\usepackage{xcolor}

\newcolumntype{?}{!{\vrule width 1pt}}

\newcommand{\nn}{f}
\newcommand{\energy}{\ensuremath{\Psi}}

\newcommand{\weights}{W}

\newcommand{\lang}{\mathcal{L}}

\newcommand{\inps}{{\mathbf x}}

\newcommand{\outs}{{\mathbf y}}

\newcommand{\reals}{\mathbb{R}}

\newcommand{\closs}{g}

\newcommand{\ncount}{\text{ct}}
\newcommand{\twonorm}[1]{\|#1\|_2}
\newcommand{\Astar}{{\texttt{A\textsuperscript{*}}}}

\DeclareMathOperator{\argmax}{argmax}

\DeclareMathOperator{\st}{s.t.}

\newcommand{\citet}[1]{\citeauthor{#1}\shortcite{#1}}
\newcommand{\citep}{\cite}

\title{Gradient-based Inference for Networks \\with Output Constraints}

\author{Jay Yoon Lee,\textsuperscript{\rm 1}\thanks{corresponding authors} \Large \textbf{Sanket Vaibhav Mehta,\textsuperscript{\rm 1} Michael Wick\textsuperscript{\rm 2}}\footnotemark[1] \\ \Large \textbf{Jean-Baptiste Tristan,\textsuperscript{\rm 2} Jaime Carbonell\textsuperscript{\rm 1}}\\ 
\textsuperscript{\rm 1}School of Computer Science, Carnegie Mellon University, Pittsburgh, PA\\
\textsuperscript{\rm 2}Oracle Labs, Burlington, MA \\
\textsuperscript{\rm 1}\{jaylee, svmehta, jgc\}@cs.cmu.edu \\
\textsuperscript{\rm 2}\{michael.wick, jean.baptiste.tristan\}@oracle.com
}
 
\begin{document}

\maketitle

\begin{abstract}
Practitioners apply neural networks to increasingly complex problems
in natural language processing, such as syntactic parsing and
semantic role labeling that
have rich output structures. Many such structured-prediction
problems require deterministic constraints on the output values; for
example, in sequence-to-sequence syntactic parsing, we require that
the sequential outputs encode valid trees. While hidden units might
capture such properties, the network is not always able to learn
such constraints from the training data alone, and practitioners
must then resort to post-processing.  In this paper, we present an
inference method for neural networks that enforces deterministic
constraints on outputs without performing rule-based post-processing
or expensive discrete search.  Instead, in the spirit of
gradient-based training, we enforce constraints with gradient-based
{\em inference} (GBI): for each input at test-time, we nudge continuous
model weights until the network's unconstrained inference procedure
generates an output that satisfies the constraints.  We study the
efficacy of GBI on three tasks with hard constraints:
semantic role labeling, syntactic parsing, and sequence transduction.  
In each case, the algorithm not only satisfies
constraints, but improves accuracy, even when the underlying network
is state-of-the-art.
\end{abstract}

\section{Introduction}
\label{sec:introduction}

Suppose we have trained a sequence-to-sequence (seq2seq) network \cite{cho14learning,sutskever14sequence,kumar16ask} to perform a structured prediction task such as syntactic constituency parsing \cite{vinyals15grammar}. We would like to apply this trained network to novel, unseen examples, but still require that the network's outputs obey an appropriate set of problem specific hard-constraints; for example, that the output sequence encodes a valid parse tree. Enforcing these constraints is important because down-stream tasks, such as relation extraction or coreference resolution typically assume that the constraints hold.  Moreover, the constraints impart informative hypothesis-limiting restrictions about joint assignments to multiple output units, and thus enforcing them holistically might cause a correct prediction for one subset of the outputs to
beneficially influence another. 

Unfortunately, there is no guarantee that the neural network will learn these constraints from the training data alone, especially if the training data volume is limited.  Although in some cases, the outputs of state-of-the-art systems mostly obey the constraints for the test-set of the data on which they are tuned, in other cases they do not. In practice, the quality of neural networks are much lower when run on data in the wild (e.g., because small shifts in domain or genre change the underlying data distribution).  In such cases, the problem of constraint violations becomes more significant.

This raises the question: how should we enforce hard constraints on the outputs of a neural network?  We could perform expensive combinatorial discrete search over a large output space, or manually construct a list of post-processing rules for the particular problem domain of interest.  Though, we might do even better if we continue to ``train'' the neural network at test-time to learn how to satisfy the constraints on each input.  Such a learning procedure is applicable at test-time because learning constraints
requires no labeled data: rather, we only require a function that
measures the extent to which a predicted output violates a constraint.

In this paper, we present \textit{gradient-based inference} (GBI), an inference method for neural networks that strongly favors respecting output constraints by adjusting the network's weights at test-time, for each input. Given an appropriate function that measures the extent of a constraint violation, we can express the hard constraints as an optimization problem over the continuous weights and apply back-propagation to tune them. That is, by iteratively adjusting the weights so that the neural network becomes increasingly likely to produce an output configuration that obeys the desired constraints. Much like scoped-learning, the algorithm customizes the weights for each example at test-time \cite{blei02learning}, but does so in a way
to satisfy the constraints.  

We study GBI on three tasks: semantic role labeling (SRL), syntactic constituency parsing and a synthetic sequence transduction problem and find
that the algorithm performs favorably on all three tasks. In summary, our contributions are that we:
\begin{enumerate}
	\item Propose a novel Gradient-Based Inference framework.
	\item Verify that GBI performs well on various applications, thus providing strong evidence for the generality of the method.
	\item Examine GBI across wide range of reference model performances and report its consistency.
	\item Show that GBI also perform well on out-of-domain data.
\end{enumerate}
For all the tasks, we find that GBI satisfies a large percentage of the constraints (up to 98\%) and that in almost every case (out-of-domain data, state-of-the art networks, and even for the lower-quality networks), enforcing the constraints improves the accuracy.  On SRL, for example, the method successfully injects truth-conveying side-information via constraints, improving SOTA network \footnote{Since our submission, the previous SOTA \cite{peters18deep} in SRL on which we apply our technique has been advanced by 1.7 F1 points ~\cite{DBLP:conf/emnlp/span-selection-OuchiS018}. However, this is a training time improvement which is orthogonal to our work.} 
by 1.03 F1 ~\cite{peters18deep}. This improvement happens to surpass a \Astar algorithm for incorporating constraints while also being robust, in a way that
\Astar is not, to cases for which the side constraints are inconsistent with the labeled ground-truth.

\section{Constraint-aware inference in neural networks}

Our goal is to design an {\em approximate} optimization algorithm that is similar in spirit to Lagrangian relaxation in that we replace a complex constrained decoding objective with a simpler unconstrained objective that we can optimize with gradient descent \cite{koo10dual,rush10dual,rush12tutorial}, but is better suited for non-linear non-convex optimization with global constraints that do not factorize over the outputs. Although the exposition in this section revolves around Lagrangian relaxation, we emphasize that the purpose is merely to provide intuition and motivate design choices.

\subsection{Problem definition and motivation}
Typically, a neural network parameterized by weights $\weights$ is a
function from an input $\inps$ to an output $\outs$. The network has
an associated compatibility function $\energy(\outs;\inps,\weights)\rightarrow\reals_+$ that measures how
likely an output $\outs$ is given an input $\inps$ under weights
$\weights$. The goal of inference is to find an output that maximizes
the compatibility function and this is usually accomplished
efficiently with feed-forward greedy-decoding. In this work, we
want to additionally enforce that the output values belong to a
feasible set or grammar $\lang^\inps$ that in general depends on the
input. We are thus interested in the following optimization problem:
\small
\begin{equation}
\label{eq:primal}
\begin{array}{ll@{}ll}
\underset{\outs}{\max}  & \displaystyle \energy(\inps,\outs,\weights) \text{      }  
\st &  \text{   } \displaystyle \outs\in\lang^\inps
\end{array}
\end{equation}
\normalsize
Simple greedy inference are no longer sufficient since the
outputs might violate the global constraints (i.e.,
$\outs\notin\lang^\inps$).  Instead, suppose we had a function
$\closs(\outs,\lang^\inps)\rightarrow\reals_+$ that measures a loss between output $\outs$ and a grammar $\lang^\inps$ such that
$\closs(\outs,\lang^\inps)=0$ if and only if there are no grammatical errors in $\outs$. That is, $\closs(\outs,\lang^\inps)=0$ for the feasible region and is strictly positive everywhere else. For example, if the feasible region is a CFL, $\closs$ could be the {\em least errors count} function \cite{lyon74syntax}.  We could then express the constraints as an equality constraint and minimize the Lagrangian:
\small
\begin{equation}
\begin{array}{ll@{}ll}
& \underset{\lambda}{\min}\;\underset{\outs}{\max}\;  & \displaystyle
\energy(\inps,\outs,\weights) + \lambda\closs(\outs,\lang^\inps)  & \\
\end{array}
\end{equation}
\normalsize
However, this leads to optimization difficulties because there is just
a single dual variable for our global constraint, resulting 
intractable problem and thus leading to
brute-force trial and error search.  

Instead, we might circumvent these issues if we optimize over a
model parameters rather than a single dual variable.
Intuitively, the purpose of the dual variables is to simply penalize
the score of {\em infeasible} outputs that otherwise have a high score in the network, but happen to violate constraints.  Similarly, network's weights can control the compatibility of the output configurations with the input. By properly adjusting the weights, we can affect the outcome of inference by removing mass from invalid outputs---in much the same way a dual variable affects the outcome of inference.  Unlike a single dual variable however, the network expresses a {\em different } penalty weight for each output. And, because the weights are typically tied across space (e.g., CNNs) or time (e.g., RNNs) the weights are likely to generalize across related outputs.  As a result, lowering the compatibility function for a single invalid output has the potential effect of lowering the compatibility for an entire family of related, invalid outputs; enabling faster search.  In the next subsection, we propose a novel approach that utilizes the amount of constraint violation as part of the objective function so that we can adjust the model parameters to search for a constraint-satisfying output efficiently.

\subsection{Algorithm}

Instead of solving the aforementioned impractical optimization
problem, we propose to optimize a ``dual'' set of model parameters
$\weights_\lambda$ over the constraint function while regularizing 
$\weights_\lambda$ 
to stay close to the originally learned
weights $W$. The objective function is as follows:
\small
\begin{equation}
\label{eq:objective}
\begin{array}{ll@{}ll}
\underset{\weights_{\lambda}}{\min} & \displaystyle \energy(\inps,\hat{\outs},\weights_\lambda)\closs(\hat{\outs},\lang^\inps) + \alpha\twonorm{\weights - \weights_\lambda} \\
\text{where} & \hat{\outs} = \displaystyle \underset{\outs}\argmax\;\energy(\inps,\outs,\weights_\lambda)
\end{array}
\end{equation}
\normalsize
Although this objective deviates from the original optimization
problem, it is reasonable because by definition of the constraint
loss $\closs(\cdot)$, the global minima must correspond to outputs
that satisfy all constraints.  Further, we expect to find
high-probability optima if we initialize $\weights_\lambda=\weights$.
Moreover, the objective is intuitive: if there is a constraint
violation in $\hat{\outs}$ then $\closs(\cdot)>0$ and the gradient
will lower the compatibility of $\hat{\outs}$ to make it less likely.
Otherwise, $\closs(\cdot)=0$ and the gradient of the energy is zero
and we leave the compatibility of $\hat{\outs}$ unchanged. Crucially,
the optimization problem yields computationally efficient subroutines
that we exploit in the optimization algorithm.

\begin{algorithm}
	\begin{algorithmic}
		\STATE{Inputs: test instance $\inps$, input specific CFL $\lang^\inps$,
			max epoch $M$, pretrained weights $\weights$}
		\STATE{$\weights_\lambda\gets\weights \,$ {\#reset instance-specific weights}} 
		\WHILE{$\closs(\outs,\lang^\inps)>0$ and iteration$<M$}
		\STATE{$\outs \gets \nn(\inps;\weights_\lambda) \,$ {\#perform inference
				using weights $\weights_\lambda$}}
		\STATE{$\nabla\gets\  \closs(\outs,\lang^\inps)\frac{\partial}{\partial\weights_\lambda}
			\energy(\inps,\outs,\weights_\lambda)
			+ \alpha\frac{\weights-\weights_\lambda}{\twonorm{\weights-\weights_\lambda}}$ {\#compute constraint loss}}
		\STATE{$\weights_\lambda \gets \weights_\lambda - \eta\nabla \,$
			{\#update instance-specific weights with SGD or a variant thereof}}
		\ENDWHILE
	\end{algorithmic}
	\caption{Constrained inference for neural nets}
	\label{alg:constraints}
\end{algorithm}

To optimize the objective, the algorithm alternates maximization to
find $\hat{\outs}$ and minimization w.r.t. $\weights_\lambda$
(Algorithm~\ref{alg:constraints}). In particular, we first approximate
the maximization step by employing the neural network's inference
procedure (e.g., greedy decoding, beam-search, or Viterbi decoding) to find the
$\hat{\outs}$ that approximately maximizes $\energy$, which ignores the
constraint loss $\closs$.  Then, given a fixed $\hat{\outs}$, we minimize the
objective with respect to the $\weights_\lambda$ by performing
stochastic gradient descent (SGD).  Since $\hat{\outs}$ is fixed, the
constraint loss term becomes a constant in the gradient; thus, making
it easier to employ external black-box constraint losses (such as
those based on compilers) that may not be differentiable.  As a
remark, note the similarity to REINFORCE \cite{williams92simple}: the decoder outputs as 
actions and the constraint-loss as 
a negative reward.  However, GBI
does not try to reduce {\em expected} reward and 
terminates upon discovery of an output that satisfies all constraints.
Furthermore, GBI also works on sequence-tagging problem, SRL (Section \ref{sec:SRL-exp}), where 
next output does not depend on the current output, which is far from REINFORCE setting.

\section{Applications}
\label{sec:application}

There are multiple applications that involve hard-constraints and we
provide two illustrative examples that we later employ as case-studies
in our experiments: SRL and syntactic parsing. The former exemplifies
a case in which external knowledge encoded as hard constraints conveys
beneficial side information to the original task of interest while the
latter studies a case in which hard constraints are inherent to the
task of interest. Finally, we briefly mention sequence transduction as
framework in which constraints may arise.  Of course, constraints may
in general arise for a variety of different reasons, depending on the
situation. We provide example-based case studies for each application in Appendix \ref{appdx:compare-GBI-const_dec}, \ref{apdx:examples}.

\subsection{Semantic Role Labeling}
As a first illustrative example, consider SRL. SRL focuses on identifying shallow semantic information about phrases. For example, in the sentence ``it is really like this, just look at the bus sign'' the goal is to tag the arguments given ``is'' as the verb predicate: ``it'' as its first argument and the prepositional phrase ``like this'' as its second argument.  Traditionally SRL is addressed as a sequence labeling problem, in which the input is the sequence of tokens and the output are BIO-encoded class labels representing both the regimentation of tokens into contiguous segments and their semantic roles.

Note that the parse tree for the sentence might provide constraints that could assist with the SRL task. In particular, each node of the parse tree represents a contiguous segment of tokens that could be a candidate for a semantic role. Therefore, we can include as side-information constraints that force the BIO-encoded class labeling to produce segments of text that each agree with some segment of text expressed by a node in the parse tree.\footnote{The ground-truth parse spans do not always agree with the SRL spans, leading to imperfect side information.}  
To continue with our example, the original SRL sequence-labeling might incorrectly label ``really like this'' as the second argument rather than ``like this.''   Since according to the parse tree ``really'' is part of the verb phrase, thus while the tree contains the spans ``is really like this'' and ``like this'' it does not contain the span ``really like this.''  The hope is that enforcing the BIO labeling to agree with the actual parse spans would benefit SRL.  Based on the experiments, this is indeed the case, and our hypothetical example is actually a real data-case from our experiments, which we describe later.  
The $\closs(\outs,\lang^x)$ for SRL factorizes into per-span constraints $\closs_i$. 
For $i$th span $s_i$, if $s_i$ is consistent with any node in the parse tree, 
$\closs_i(s_i,\lang^x)=0$, otherwise $\closs_i(s_i,\lang^x)=1/n_{s_i}$ where $n_{s_i}$ is defined as the number of tokens in $s_i$.
Overall,
$\energy(\inps,\hat{\outs},\weights_\lambda)\closs(\hat{\outs},\lang^x)=
\sum_{i=1}^k \closs(s_i,\lang^x)\energy(\inps,s_i,W_\lambda)$ where $k$ is number of spans on output $\hat{\outs}$.

\subsection{Syntactic parsing}

\label{subsec:parsing}
As a second illustrative example, consider a structured prediction problem 
of syntactic parsing in which the goal is to input a sentence
comprising a sequence of tokens and output a tree describing the
grammatical parse of the sentence. 
Syntactic parsing is a separate but complementary task to SRL. 
While SRL focuses on semantic information, syntactic parsing focuses on identifying relatively deep syntax tree structures.
One way to model the problem with
neural networks is to linearize the representation of the parse tree
and then employ the familiar seq2seq model
\cite{vinyals15grammar}.  Let us suppose we linearize the tree using
a sequence of shift (\texttt{s}) and reduce (\texttt{r,r!}) commands
that control an implicit shift reduce parser. Intuitively, these
commands describe the exact instructions for converting the input
sentence into a complete parse tree: the interpretation of the symbol
\texttt{s} is that we shift an input token onto the stack and the
interpretation of the symbol \texttt{r} is that we start (or continue)
reducing (popping) the top elements of the stack, the interpretation
of a third symbol \texttt{!}  is that we stop reducing and push the
reduced result back onto the stack. Thus, given an input sentence and
an output sequence of shift-reduce commands, we can deterministically
recover the tree by simulating a shift reduce parser. For example, the
sequence \texttt{ssrr!ssr!rr!rr!} encodes a type-free version of the
parse tree \texttt{(S (NP the ball) (VP is (NP red)))} for the input
sentence ``the ball is red''. It is easy to recover the tree structure
from the input sentence and the output commands by simulating the shift
reduce parser.  Of course in practice, reduce commands
include the standard parts of speech as types (NP, VP, etc).

Note that for output sequences to form a valid tree over the input,
the sequence must satisfy a number of constraints. First, the number
of shifts must equal the number of input tokens $m_\inps$, otherwise
either the tree would not cover the entire input sentence or the tree
must contain spurious symbols. Second, the parser cannot
issue a reduce command if the stack is empty. Third, 
at the end of the parser commands,
the stack must have just a single item,
the root node. 
The constraint loss $\closs(\outs,\lang^x)$ for this task simply counts the errors of each
of the three types. (Appendix \ref{appdx:parse-constraint})

As a minor remark, note that other encodings of trees, such as
bracketing (of which the Penn Tree Bank's S-expressions are an
example), are more commonly used as output representations for seq2seq
parsing ({\em ibid}).  However, the shift-reduce representation
described in the above paragraphs is isomorphic to the bracketing
representations and as we get similar model performance to single
seq2seq mode  on the same data ({\em ibid.}), we chose the former
representation to facilitate constraint analysis.  Although output
representations sometimes matter, for example, BIO vs BILOU encoding
of sequence labelings, the difference is usually minor
\cite{ratinov09design}, and breakthroughs in sequence labeling have
	been perennially advanced under both representations.
Thus, for now, we embrace the shift reduce representation as a legitimate alternative to bracketing, {\em pari passu}.

\subsection{Synthetic sequence transduction}
Finally, although not a specific application per se, we also consider sequence transduction as it provides a framework conducive to studying simple artificial languages with appropriately designed properties.  A sequence transducer $T:\lang_S\rightarrow\lang_T$ is a function from a source sequence to a target sequence.  As done in previous work, we consider a known $T$ to generate input/output training examples and train a seq2seq network to learn $T$ on that data \cite{grefenstette15learning}.  
The constraint is simply that the output must belong to $\lang_T$ and also respect problem-specific conditions that may arise from the application of $T$ on the input sentence. 
We study a simple case in Section~\ref{subsec:exp-transducer}.

\section{Experiments}
\label{sec:experiments}

\begin{table*}
	\small
	\centering
	\begin{tabular}{|c|c|c|c|c|r|r|r|c|r|r|r|}
		\hline
		\multirow{4}{*}{Network} & \multirow{4}{*}{Failure} & \multirow{4}{*}{Inference} & \multirow{4}{*}{Conv} & \multicolumn{6}{|c|}{Failure set} & \multicolumn{2}{c|}{Test set}\\
		\cline{5-12}
		& & & & \multicolumn{2}{|c|}{Average (\%)} &  \multicolumn{2}{c|}{\multirow{2}{*}{F1} } & \multicolumn{2}{c|}{Exact Match} & \multicolumn{2}{c|}{\multirow{2}{*}{F1}}\\
		& \multirow{2}{*}{rate(\%)} & & \multirow{2}{*}{rate(\%)} & \multicolumn{2}{|c|}{Disagreement} &  \multicolumn{2}{c|}{} & \multicolumn{2}{c|}{(\%)} & \multicolumn{2}{c|}{ }\\
		
		\cline{5-12}
		& & & & before & after &  before & \multicolumn{1}{c|}{after} & before & after & before & \multicolumn{1}{c|}{after} \\
		\hline
		\multirow{2}{*}{SRL-100} & \multirow{2}{*}{9.82} & GBI  & 42.25 & \multirow{2}{*}{44.85} & 24.92 & \multirow{2}{*}{48.00} & 59.70 (+11.7) & \multirow{2}{*}{0.0} & 19.90 & \multirow{2}{*}{84.40} & 85.63 (+1.23) \\
		& & \Astar & 40.40 &  & 33.91 &  & 48.83 (+0.83) & & 13.79 &  & 84.51 (+0.11) \\
		\hline
		\multirow{2}{*}{SRL-70}  & \multirow{2}{*}{10.54} & GBI  & 46.22 & \multirow{2}{*}{45.54} & 23.02 & \multirow{2}{*}{47.81} & 59.37 (+11.56) & \multirow{2}{*}{0.0} & 19.57 & \multirow{2}{*}{83.55} & 84.83 (+1.28)\\
		& & \Astar & 44.42 &  & 32.32 &  & 50.49 (+2.68) & & 16.12 &  & 83.90 (+0.35)\\
		\hline
		\multirow{2}{*}{SRL-40}  & \multirow{2}{*}{11.06} & GBI & 47.89 & \multirow{2}{*}{45.71} & 22.42 & \multirow{2}{*}{46.53} & 58.83 (+12.3) & \multirow{2}{*}{0.0} & 19.45 & \multirow{2}{*}{82.57} & 84.03 (+1.46)\\ 
		& & \Astar & 44.74 &  & 32.17 &  & 46.53 (+2.88) &  & 15.15 &  & 82.98 (+0.41) \\
		\hline             
		\multirow{2}{*}{SRL-10}  & \multirow{2}{*}{14.15} & GBI  & 44.28 & \multirow{2}{*}{47.14} & 24.88 & \multirow{2}{*}{44.19} & 54.78 (+10.59) & \multirow{2}{*}{0.0} & 15.28 & \multirow{2}{*}{78.56} & 80.18 (+1.62) \\
		& & \Astar & 43.66 &  & 32.80 &  & 45.93 (+1.74) &  & 12.28 &  & 78.87 (+0.31)\\
		\hline
		\multirow{2}{*}{SRL-1}  & \multirow{2}{*}{21.90} & GBI  & 52.85 & \multirow{2}{*}{50.38} & 21.45 & \multirow{2}{*}{37.90} & 49.00 (+11.10) & \multirow{2}{*}{0.0} & 12.83 & \multirow{2}{*}{67.28} & 69.97 (+2.69) \\
		& & \Astar & 48.96 &  & 30.28 &  & 41.59 (+3.69) & & 11.25 &  & 67.97 (+0.69)\\  \hline
	\end{tabular}
	\caption{Comparison of the GBI vs. \Astar inference procedure
			for SRL.  We report the avg. disagreement rate, F1-scores and exact match for the {\em failure set} (columns 5-10) and F1-score for the whole test set (last 2 columns). Also, we report performances on a wide range of reference models SRL-X, where X denotes \% of dataset used for training. We employ Viterbi decoding as a base inference strategy (before) and apply GBI (after) in combination with Viterbi.}
	\label{tab:srl}
\end{table*}

\begin{table}
	\small
	\begin{center}
		\begin{tabular}{|c|c|c|c|c|c|c|}
			\hline
			\multirow{2}{*}{name} & \multicolumn{2}{c}{F1} & \multicolumn{3}{|c|}{hyper-parameters} & data \\
			\cline{2-6}
			& BS-9 & greedy & hidden & layer & dropout & (\%) \\
			\hline
			\texttt{Net1} & 87.58 & 87.31 & 128 & 3 & 0.5 & 100 \\
			\texttt{Net2} & 86.63 & 86.54 & 128 & 3 & 0.2 & 100 \\
			\hline
			\hline
			\texttt{Net3} & 81.26 & 78.32 & 172 & 3 & no  & 100 \\
			\texttt{Net4} & 78.14 & 74.53 & 128 & 3 & no & 75 \\
			\texttt{Net5} & 71.54 & 67.80 & 128 & 3 & no & 25 \\
			\hline
		\end{tabular}
		\caption{Parsing Networks with various performances (BS-9 means beam size 9). \texttt{Net1,2} are GNMT seq2seq models whereas \texttt{Net3-5} are lower-resource and simpler seq2seq models, providing a wide range of model performances on which to test GBI.}
		\label{tab:parsing_networks}
	\end{center}
\end{table}

\begin{table}[h]
	\small
	\centering
	\begin{tabular}{|c|r|r|r|r|r|r|} 
		\hline
		\multirow{2}{*}{Net} &  Failure  & Conv & \multicolumn{2}{c|}{F1 (Failure set) } & \multicolumn{2}{c|}{F1 (whole test) } \\
		\cline{4-7}
		& (/2415)      & rate         & before & after & before & after \\
		\hline
		Net1 & 187 &  93.58 &  71.49 &  77.04  & 87.31 & 87.93 \\
		Net2 & 287 &  89.20 &  73.54 &  79.68  & 86.54 & 87.57 \\
		\hline
	\end{tabular}
	\caption{Evaluation of GBI on syntactic parsing using GNMT seq2seq. Note that GBI without beam search performs higher than BS-9 in Table \ref{tab:parsing_networks}.}
	\label{tab:parsing_best}
\end{table}

In this section we study our algorithm on three different tasks: SRL, syntactic constituency parsing and a synthetic sequence transduction task.
All tasks require hard constraints, but they play a different role in each.  
In the transduction task they force the
output to belong to a particular input-dependent regular expression, 
in SRL, constraints provide side-information about possible
true-spans and in parsing,
constraints ensure that the outputs encode valid trees.
While the SRL task involves more 
traditional recurrent neural networks that have 
exactly one output per input token,
the parsing and transduction tasks 
provide an opportunity to study the algorithm 
on various seq2seq networks .

We are interested in answering the following questions (Q1) how well
does the neural network learn the constraints from data (Q2) for cases
in which the network is unable to learn the constraints perfectly,
can GBI actually enforce the constraints 
(Q3) does GBI enforce constraints without compromising the quality of the
network's output.  To more thoroughly investigate Q2 and Q3, we also
consider: (Q4) is the behavior of the method sensitive to the
reference network performance,
and (Q5) does GBI also work on out-of-domain data.
Q3 is particularly important because we adjust the weights of
the network at test-time and this may lead to unexpected behavior.
Q5 deals with our original motivation of using structured prediction 
to enhance performance on the out-of-domain data.

To address these various questions, we first define some terminology to measure 
how well the model is doing in terms of constraints.
To address (Q1) we
measure the {\em failure-rate} (i.e., the ratio of test sentences for
which the network infers an output that fails to fully satisfy the
constraints).  To address (Q2) we evaluate our method on the
{\em failure-set} (i.e., the set of output sentences for which the
original network produces constraint-violating outputs) and
measure our method's {\em conversion rate}; that is, the percentage of
failures for which our method is able to completely satisfy the
constraints (or ``convert'').  Finally, to address (Q3), we evaluate
the quality (e.g., accuracy or F1) of the output predictions on the
network's {\em failure-set} both before and after applying our method.

\subsection{Semantic Role Labeling}
\label{sec:SRL-exp}

We employ the AllenNLP \cite{Gardner2017AllenNLP} SRL network
with ELMo embeddings, which is a multi-layer highway
bi-LSTM that produces BIO output predictions for each input token
\cite{peters18deep}. For data we use OntoNotes v5.0, which has
ground-truth for both SRL and syntactic parsing
\cite{pradhan2013towards}. We evaluate GBI on the test-set
(25.6k examples), out of which consistent parse information is available for
81.25\% examples (we only include side-information in terms of constraints for this subset).

We repeat the same experimental procedure over multiple networks,
SRL-X, while varying the portion (X\%) of the training dataset. 
In Table \ref{tab:srl}, we see that GBI is able to convert 42.25 \% of failure set, and this boosts
the overall F1 measure by 1.23 point over the SOTA network (SRL-100) which does not incorporate the
constraints (they report 84.6 F1, we obtain a similar 84.4 F1 with their
network, and achieve 85.63 after enforcing constraints with our
inference). Further, to address (Q1) we measure the sentence-level {\it failure rate} as well as span-level
\textit{disagreement rate} (i.e., the ratio of predicted spans in
a sentence that disagree with the spans implied by the true syntactic
parse of the sentence). To address (Q2) we evaluate our method on the
\textit{failure set} (i.e., the set of sentences for which
disagreement rate is nonzero) and measure our method's
avgerage disagreement rate. Finally, to address (Q3), we evaluate the
quality (F1 and exact match) of the output predictions on the network's
\textit{failure-set} both before and after applying our method.  From
Table \ref{tab:srl}, we can see that by applying GBI on SRL-100,
the avgerage disagreement rate on the failure set goes down
from 44.85\% to 24.92\% which results in an
improvement of 11.7 F1 and 19.90\% in terms of exact match on the same 
set. 
These improvements answer Q1-3 favorably.

To enforce constraints during inference, \citeauthor{he2017deep}
proposed to employ constrained-\Astar decoding. For the sake of a
fair comparison with GBI, we consider \Astar decoding as used in
\cite{he2017deep} and report results for the SRL-X networks. We see
from Table \ref{tab:srl}, that the GBI procedure consistently
outperforms \Astar decoding on all evaluation metrics, thus
demonstrating the superiority of the approach.

\begin{table}
	\small
	\centering
	\begin{tabular}{|c|c|r|r|r|r|} 
		\hline
		\multirow{2}{*}{Net} & Infer & \multicolumn{1}{c|}{Failure}  & \multicolumn{1}{c|}{Conv} & \multicolumn{2}{c|}{F1 (Failure set)}\\
		\cline{5-6}
		& method & (/2415) & \multicolumn{1}{c|}{rate} & before & \multicolumn{1}{c|}{after}\\
		\hline
		\multirow{4}{*}{Net3} 
		& Greedy & 317 &  79.81  &  65.62 &  68.79 (+3.14)\\
		& Beam 2 & 206 & 87.38  &  66.61 &  71.15 (+4.54)\\
		& Beam 5 & 160 & 87.50  &  67.5 &  71.38 (+3.88)\\
		& Beam 9 & 153 & 91.50  & 68.66  &  71.69 (+3.03)\\
		\hline
		\multirow{4}{*}{Net4} 
		& Greedy & 611 &  88.05 & 62.17  & 64.49 (+2.32)\\
		& Beam 2 & 419 & 94.27 & 65.40  & 66.65  (+1.25)\\
		& Beam 5 & 368  &  92.66 & 67.18 & 69.4  (+2.22)\\
		& Beam 9 & 360  &  93.89 & 67.83 & 70.64 (+2.81)\\
		\hline
		\multirow{4}{*}{Net5} 
		& Greedy & 886 &  69.86 &  58.47 &  60.41 (+1.94)\\
		& Beam 2 & 602 & 82.89 &  60.45 &  61.35  (+0.90)\\
		& Beam 5 & 546 &  81.50 &  61.43 &  63.25 (+1.82) \\
		& Beam 9 & 552 &  80.62 &  61.64 &  62.98 (+1.34) \\
		\hline
	\end{tabular}
	\caption{Evaluation of GBI on simpler, low-resource seq2seq networks. Here, we also evaluate whether GBI can be used in combination with different inference techniques: greedy and beam search of various widths.}
	\label{tab:parsing_low}
\end{table}

\subsection{Syntactic parsing}
We now turn to a different task and network: syntactic constituency parsing.
We investigate the behavior of the constraint inference algorithm
on the shift-reduce parsing task described in
Section~\ref{sec:application}. We transform the Wall Street Journal (WSJ)
portion of the Penn Tree Bank (PTB) into shift-reduce commands in
which each reduce command has a phrase-type (e.g., noun-phrase or
verb-phrase) \cite{mitchell99ptb}. We employ the traditional split of
the data with section 22 for dev, section 23 for test, and remaining
sections 01-21 for training. We evaluate on the test set with
evalb\footnote{\url{http://nlp.cs.nyu.edu/evalb/}} F1. In each experiment, we learn a seq2seq network on a
training set and then evaluate the network directly on the test set
using a traditional inference algorithm to perform the decoding
(either greedy decoding or beam-search).  

In order to study our algorithm on a wide range of 
accuracy regimes (section \ref{subsec:sensitivity}), 
we train many networks with different
hyper-parameters producing models of various quality, from high to
low, using the standard split of the WSJ portion of the PTB. 
In total, we train five networks \texttt{Net1-5} for this study, that we
describe below. We train our two best baseline models ({\texttt Net1,2}) using a
highly competitive seq2seq architecture for machine
translation, GNMT \cite{wu16google} with F1 scores, 86.78 and 87.33, respectively.  And, to study a wider range of accuracies, we train a simpler architecture with different hyper parameters and
obtain nets ({Net3-5}). For all models, we employ Glorot initialization, and basic attention \cite{bahdanau14neural}. 
See Table~\ref{tab:parsing_networks} for a summary of the networks,
hyper-parameters, and their performance.

We report the behavior of the constraint-satisfaction method 
in Table~\ref{tab:parsing_best} for Net1-2, 
and in Table~\ref{tab:parsing_low} for Net3-5. 
Across all the
experimental conditions (Table \ref{tab:parsing_best}, \ref{tab:parsing_low}), the conversion rates are high, often above 80
and sometimes above 90 supporting Q2. 
Note that beam search alone can also increase constraint satisfaction
with conversion rates reaching as high as 51.74\% (164/317) in the
case of Net3 with beam size 9. However, as the quality of the model
increases, the conversion rate becomes minuscule; in the case of
Net1,2 the conversion rate is less than 14\% with beam 9; in Net1
converting 26 out of 187 and in Net2 converting just 1 out of 287 instances from failure set.

In order to address question Q3---the ability of our approach to
satisfy constraints without negatively affecting output quality---we
measure the F1 scores on the failure-sets both before and after
applying the constraint satisfaction algorithm.
Since F1 is only defined on valid trees, 
	we employ heuristic post-processing to ensure all outputs are valid.

Note that an improvement on the failure-set
guarantees an improvement on the entire test-set since our method
produces the exact same outputs as the baseline for examples that do
not initially violate any constraints.  
Consequently, for example, the
GNMT network improves (Net2) on the failure-set from 73.54 to 79.68
F1, resulting in an overall improvement from 86.54 to 87.57 F1  (entire test-set).  
These improvements are similar to those we observe in the SRL task, and 
provide additional evidence for answering Q1-3 favorably.
We also measure how many iterations 
of our algorithm 
it takes to
convert the examples that have constraint-violations.  Across all
conditions, it takes 5--7 steps to convert 25\% of the outputs, 6--20
steps to convert 50\%, 15--57 steps to convert 80\%, and 55--84 steps to convert 95\%.

\subsection{Simple Transduction Experiment}
\label{subsec:exp-transducer}
In our final experiment we focus on a simple sequence transduction
task in which we find that despite learning the training data
perfectly, the network fails to learn the constraint in a way that
generalizes to the test set.  

For our task, we choose a simple transducer, similar to those studied
in recent work \cite{grefenstette15learning}. The source language
$\lang_S$ is \texttt{(az|bz)}$^\star$ and the target language
$\lang_T$ is \texttt{(aaa|zb)}$^\star$. The transducer is defined to
map occurrences of \texttt{az} in the source string to \texttt{aaa} in
the target string, and occurrences of \texttt{bz} in the source string
to \texttt{zb} in the target string. For example,
$T(\texttt{bzazbz})\mapsto\texttt{zbaaazb}$.  The training set
comprises 1934 sequences of length 2--20 and the test set contain
sentences of lengths 21-24. We employ shorter sentences for training
to require generalization to longer sentences at test time.

We employ a 
32 hidden unit single-layered, attention-less,
seq2seq LSTM in which the
decoder LSTM inputs the final encoder state at each decoder time-step.
The network achieves perfect train accuracy while learning the rules of
the target grammar $\lang_T$ perfectly, even on the test-set.
However, the network fails to learn the input-specific constraint that
the number of \texttt{a}'s in the output should be three times the
number of \texttt{a}'s in the input.  This illustrates how a network
might rote-memorize constraints rather than learn the rule in a way
that generalizes. Thus, enforcing constraints at test-time is
important.  
To satisfy constraints, we employ GBI with a
constraint loss $\closs$, a length-normalized quadratic
$(3x_a-y_a)^2/(m+n)$ that is zero when the number of \texttt{a}'s in
the output ($y_a$) is exactly three times the number in the input
($x_a$) with $m$,$n$ denoting input, output, respectively.  GBI achieves a conversion rate of
65.2\% after 100 iterations, while also improving the accuracy on the
failure-set from 75.2\% to 82.4\%.  This synthetic experiment provides
additional evidence in support of Q2 and Q3, on a simpler
small-capacity network.

\begin{table*}
	\small
	\centering
	\begin{tabular}{|c? r|r|r|c ? r|r|r|c|} 
		\hline
		& \multicolumn{4}{c?}{ \bf \em Syntactic Parsing} & \multicolumn{4}{c|}{\bf \em  SRL} \\
		\cline{2-9}
		Genre & Failure & Conversion  & \multicolumn{2}{c?}{F1 on failure set} &  Failure & Conversion  & \multicolumn{2}{c|}{F1 on failure set}   \\
		\cline{4-5}\cline{8-9}
		& rate (\%) & rate (\%) & before & after & rate (\%) & rate (\%) & before & after\\
		\hline
		Broadcast Conversation (BC)  & 19.3 & 98.8 & 56.4 & 59.0 (+2.6) & 26.86 & 53.88 & 39.72 & 52.4 (+12.68)\\
		Broadcast News (BN) & 11.7 & 98.1 & 63.2 & 68.8  (+5.6) & 18.51 & 55.19 & 39.28 & 50.58 (+11.3)\\
		Pivot Corpus (PT) & 9.8  & 97.8 & 71.4 & 75.8  (+4.4) & 10.01 & 62.34 & 47.19 & 63.69 (+16.5)\\
		Telephone Conversation (TC) & 10.1 & 86.2 & 56.9 & 57.6. (+0.7) & 19.09 & 54.62 & 47.7 & 58.04 (+10.34)\\
		Weblogs (WB) & 17.6 & 95.3 & 62.0 & 63.2  (+1.2) & 20.32 & 44.13 & 47.6 & 57.39 (+9.39)\\
		\hline
	\end{tabular}
	\caption{Evaluation of syntactic parser and SRL system on out-of-domain data. 
		F1 scores are reported on the {\em failure set}. 
		SRL model was trained on  NW and the syntactic parser was trained on WSJ Section on OntoNote v5.0.
		Except PT, which is new and old Testament, all failure rate on out-domain data is higher than that of in-domain (11.9\% for parsing and 18.1\% for SRL) 
		as suspected. 
		The table shows that GBI can be successfully applied to resolve performance degradation on out-of-domain data.
	}
	\label{tab:out-of-genre}
\end{table*}

\subsection{ GBI on wide range of reference models}
\label{subsec:sensitivity}
The foregoing experimental results provide evidence that GBI is a viable method for enforcing constraints.  However, we hitherto study GBI on high quality reference networks such as SRL-100.  To further bolster our conclusions, we now direct our investigation towards lower quality networks to understand GBI's viability under a broader quality spectrum. We ask, how sensitive is GBI to the reference network's performance (Q4)? To this end, we train poorer quality networks by restricting the amount of available training data or employing simpler architectures.

For {\em SRL}, we simulate low-resource models by limiting the training data portion to 1\%, 10\%, 40\%, and 70\% resulting in F1 score range of 67.28-83.55.
Similarly, for {\em syntactic parsing}, we train additional low-quality models {\texttt Net3-5} with a simpler uni-directional encoders/decoders, and on different training data portions of 25\%, 75\%, and 100\% (Table \ref{tab:parsing_networks}).  
We evaluate GBI on each of them in Table~\ref{tab:srl},~\ref{tab:parsing_low} and find further evidence in support of favorable answers to 
Q2 (satisfying constraints) and Q3 (improving F1 accuracy) by favorably answering Q4.
Moreover, while not reported fully due to page limits, we examined both tasks with over 20 experiments and different baseline networks in combination with different inference strategies, and we found GBI favorable in all but one case (but by just 0.04 comparing without GBI).

We also study whether GBI is compatible with better underlying discrete search algorithms, in particular beam search for seq2seq.  As we seen in column~2 of Table~\ref{tab:parsing_low}, that although beam-search improves the F1 score and reduces the percentage of violating constraints, GBI further improves over beam-search when using the latter in the inner-loop as the decoding procedure.  In conclusion, improving the underlying inference procedure has the effect of decreasing the number of violating outputs, but GBI is still very much effective on this increasingly small set, despite it intuitively representing more difficult cases that even eludes constraint satisfaction via beam search inference.

\subsection{ Experiments on out-of-domain data} 
\label{subsec:out-of-dom}
Previously, we saw how GBI performs well even when the underlying network is of lower quality.  
We now investigate GBI on actual out-of-domain data for which the model quality can suffer.
For SRL, we train a SOTA network with ELMo embedding on the
	NewsWire (NW) section of the OntoNotes v5.0 English PropBank corpus and then test on the other genres provided in the corpus: 
    BC, BN, PT, TC, WB.
	The failure rate on the within genre data (test set of NW) is
	18.10\%. We can see from Table \ref{tab:out-of-genre}, the failure
	rate for the NW trained SRL network in general is higher for out-of-genre data with the highest being 26.86\% for BC (vs. 18.10\% NW). Further, by enforcing constraints, we see significant gains on the failure set in terms of F1 score across all genres (ranging from 9.39-16.5 F1), thus, providing additional evidences for answering Q5.

As we did for SRL, we train a GMNT seq2seq model on the WSJ NW section in OntoNotes v5.0 Treebank \footnote{The PTB (40k instances) and OntoNotes (30k instances) coverage of WSJ are slightly different.} which shares the same genre classification with PropBank.
The F1 on the within-genre data (test set of WSJ) is 85.03, but the F1 on 
these genres is much lower, ranging from the mid-forties on BC (46.2--78.5 depending on the subcategory) 
to the low-eighties on BN (68.3--81.3. depending on the subcategory). 
Indeed, we find that overall the F1 is lower and in some cases, like WB, the failure rate is much higher (17.6\% for WB {\em vs.} 11.9\% for WSJ).  Following the same experimental protocol as on the PTB data, we report the results in Table~\ref{tab:out-of-genre} (aggregating over all subcategories in each genre).  We see that across all genres, the algorithm has high conversion rates (sometimes close to 100\%), and that in each case, enforcing the constraints improves the F1.  Again, we find support for Q2, Q3 and Q5.

\subsection{Robustness and Runtime analysis}
We perform additional experiments to analyze the robustness and
runtime of GBI.  First, to measure robustness, we consider a variant
of the SRL task in which we include noisy constraints, and compare GBI
to \Astar (Appendix \ref{apdix:noisySRL}).  We find that in this case,
\Astar performs significantly worse than the baseline, while GBI improves over the same baseline, thus showing the robustness of GBI.

In terms of runtime, GBI is generally faster than \Astar, though, the
difference is less clear on smaller evaluation sets (Appendix \ref{apdix:runtimeSRL}). 
In the case study with noisy constraints, the runtimes are similar;
however, GBI has much better accuracy, showing similar gains as the
noise-free setting.  Lastly, in appendix \ref{apdix:runtimeSRL}, we
discuss GBI's trade off between runtime and accuracy by varying the max epoch $M$.

\section{Related work}
\label{sec:related}
Recent work has considered applying neural networks to structured prediction; for example,
structured prediction energy networks (SPENs) \cite{belanger16structured}.  SPENs incorporate soft-constraints via back-propagating an energy function into ``relaxed'' output variables.
In contrast, we focus on hard-constraints and back-propagate into the weights that subsequently control the original non-relaxed output variables via inference.  Separately, there has been interest in employing hard constraints to harness unlabeled data in
training-time for simple classifications \cite{hu16harnessing}.  
Our work instead focuses on enforcing constraints at inference-time. More specifically, for SRL, previous work for enforcing 
constraints have focused on constrained \Astar decoding \cite{he2017deep} or integer linear programming \cite{punyakanok2008importance}. For parsing, previous work in enforcing hard constraints has focused on post-processing \cite{vinyals15grammar} or building them into the decoder transitions \cite{dyer16recurrent} 
or search constraints \cite{wiseman16sequence}. 

Finally, as previously mentioned, our method highly resembles dual
decomposition and more generally Lagrangian relaxation for structured
prediction \cite{koo10dual,rush10dual,rush12tutorial}. In such
techniques, it is assumed that a computationally efficient inference
algorithm can maximize over a superset of the feasible region (this
assumption parallels our case because unconstrained inference in the
neural network is efficient, but might violate constraints). Then, the
method employs gradient descent to concentrate this superset onto the
feasible region.  However, these techniques are not directly
applicable to our non-linear problem with global constraints.

\section{Conclusion}
We presented an algorithm for satisfying constraints in neural
networks that avoids combinatorial search, but employs the network's
efficient unconstrained procedure as a black box to coax weights 
towards well-formed outputs. We evaluated the
algorithm on three tasks including SOTA SRL, seq2seq parsing and
found that GBI can successfully convert failure sets while also boosting the task performance.
Accuracy in each of the three tasks was improved by respecting constraints. Additionally, for SRL, we employed GBI on a model trained with similar constraint enforcing loss as GBI's ~\cite{srl_ssl},
and observe that the additional test-time optimization of GBI still significantly improves the model output whereas \Astar does not.
We believe this is because GBI searches in the proximity of the
provided model weights; however, theoretical analysis of this hypothesis is left as a future work.

\bibliography{ref}
\bibliographystyle{aaai}

\newpage
\onecolumn

\section*{Appendix}
\appendix
\section{GBI vs. Constrained decoding}

In Table~\ref{tab:sr-improves} of Appendix \ref{apdx:examples}, we
provide an example data-case that shows how our algorithm solves the
initially violated shift-reduce parse output.  For simplicity we
omit the phrase-types of constituency parsing and display only on the shift (\texttt{s}),
reduce (\texttt{r}) and stop reducing commands (\texttt{!}), and
color them red if there is an error.  The algorithm satisfies the
constraint in just 12 iterations, and this results in a perfectly
correct parse.  What is interesting about this example is that the
original network commits a parsing mistake early in the output
sequence.  This type of error is problematic for a naive decoder
that greedily enforces constraints at each time-step.  The reason is
that the early mistake does not create a constraint violation until
it is too late, at which point errors have already propagated to
future time-steps and the greedy decoder must shift and reduce the
last token into the current tree, creating additional spurious parse
structures.  In contrast, our method treats the constraints
holistically, and uses it to correct the error made at the beginning
of the parse. See Table~\ref{tab:sr-compare} for a comparison of
how the methods fix the constraints.  Specifically, the
constraint violation is that there were not enough shift and reduce
commands to account for all the tokens in the sentence.  Rather than
fixing the constraint by inserting these extra commands at the end
of the sequence as the greedy decoder must do, GBI inserts them at
the beginning of the sequence where the initial mistake was made,
thereby correcting the initial mistake.  Moreover, this correction
propagates to a mistake made later in the sequence (viz., the the
sequence of three reduces after the four shifts) and fixes them
too.  This example provides evidence that GBI can indeed enforce
constraints holistically and that doing so improves the output in a
global sense.
\label{appdx:compare-GBI-const_dec}
\begin{table*}[ht]
	\centering
	\begin{tabular}{|r|l|}
		\multicolumn{2}{c}{$\left<\text{`` So it 's a very mixed bag . ''}\right>$ $\longrightarrow$
			\texttt{sssr!ssssrr!srrr!rr!ssrrrrrr!}}\\
		\hline
		\hline
		inference method & output \\
		\hline
		unconstrained-decoder & \texttt{ssr!sr!ssssrrr!~~~~rr!ssrrrrrr!} \\
		constrained-decoder & \texttt{ssr!sr!ssssrrr!~~~~rr!ssrrrrrr!srr!} \\
		our method & \texttt{~~sssr!ssssrr!srrr!rr!ssrrrrrr!} \\
		true parse & \texttt{~~sssr!ssssrr!srrr!rr!ssrrrrrr!} \\
		\hline 
	\end{tabular}
	\caption{A shift-reduce example for which the method successfully
		enforces constraints. The initial unconstrained decoder prematurely
		reduces ``So it'' into a phrase, missing the contracted verb ``is.''
		Errors then propagate through the sequence culminating in the final
		token missing from the tree (a constraint violation).  The constrained decoder is only able to
		deal with this at the end of the sequence, while our method is able to
		harness the constraint to correct the early errors.
	}
	\label{tab:sr-compare}
\end{table*}

\section{Example-based case study}
\label{apdx:examples}
\begin{table*}[h]
	\centering
	\begin{tabular}{r|l|r|r}
		\multicolumn{4}{c}{$\left<\text{`` it is really like this , just look at the bus signs . ''}\right>$ $\longrightarrow$
			\texttt{B-ARG1 B-V B-ARGM-ADV B-ARG2 I-ARG2 O O $\cdots$ O}}\\
		\hline
		\hline
		iteration & output & loss & accuracy \\
		\hline
		0 & B-ARG1 B-V {\color{red}B-ARG2 I-ARG2 I-ARG2} O O O O O O O O & 0.012 & 50.0\%\\
		6 & B-ARG1 B-V {\color{red}B-ARG2 I-ARG2 I-ARG2} O O O O O O O O & 0.049 & 50.0\% \\
		7 & B-ARG1 B-V B-ARGM-ADV B-ARG2 I-ARG2 O O O O O O O O & 0.00 & 100.0\% \\
		\hline
	\end{tabular}
	\caption{A semantic role labeling example for which the method successfully
		enforces syntactic constraints. The initial output has
		an inconsistent span for token "really like this". Enforcing the constraint not only corrects the number of agreeing spans, but also changes the semantic role "B-ARG2" to "B-ARGM-ADV" and "I-ARG2" to "B-ARG2"..}
	\label{tab:srl-improves}
\end{table*}

\begin{table*}[ht]
	\centering
	\begin{tabular}{r|l|r|r}
		\multicolumn{4}{c}{$\left<\text{`` So it 's a very mixed bag . ''}\right>$ $\longrightarrow$
			\texttt{sssr!ssssrr!srrr!rr!ssrrrrrr!}}\\
		\hline
		\hline
		iteration & output & loss & accuracy \\
		\hline
		0 & \texttt{ss{\color{red}r!sr!}ss{\color{red}ssrr}r{\color{red}!}r{\color{red}r!ssrr}rrrr{\color{red}!}}  & 0.0857 & 33.3\% \\
		11 & \texttt{ss{\color{red}r!sr!}ss{\color{red}ssrr}r{\color{red}!}r{\color{red}r!ssrr}rrrr{\color{red}!}} & 0.0855 & 33.3\% \\
		12 & \texttt{sssr!ssssrr!srrr!rr!ssrrrrrr!} & 0.0000 & 100.0\% \\ 
		\hline 
	\end{tabular}	
	\caption{A shift-reduce example for which the method successfully
		enforces constraints. The initial output has only nine shifts, but there
		are ten tokens in the input. Enforcing the
		constraint not only corrects the number of shifts to ten, but
		changes the implied tree structure to the correct tree.}
	\label{tab:sr-improves}
\end{table*}

\begin{table*}[h]
	\centering
	\begin{tabular}{r|l|r|r}
		\multicolumn{4}{c}{\texttt{azazbzazbzbzazbzbzbzbzbz} $\longrightarrow$
			\texttt{aaaaaazbaaazbzbaaazbzbzbzbzb}}\\
		\hline
		\hline
		iteration & output & loss & accuracy \\
		\hline
		0 & \texttt{aaaaaazbaaazb{\color{red}aa}a{\color{red}zb}zbzbzb{\color{red}aaazb}} & 0.2472 & 66.7\\
		1 & \texttt{aaaaaazbaaazb{\color{red}aa}a{\color{red}zb}zbzbzb{\color{red}aaazb}} & 0.2467 & 66.7\\
		2 & \texttt{aaaaaazbaaazb{\color{red}aa}a{\color{red}zb}zbzbzb{\color{red}aaazb}} & 0.2462 & 66.7\\
		3 & \texttt{aaaaaazbaaazbzbaaazbzbzbzbzb} & 0.0 & 100.0 \\
		\hline
	\end{tabular}
	\caption{A sequence transduction example for which enforcing the constraints improves
		accuracy. Red indicates errors.}
	\label{tab:azbz-multi} 
\end{table*}

\newpage

\section{Constraint functions}

Here we define the specific constraint loss function
$\closs(\outs,\lang^x)$ for each task.  Note that a common theme is
that we normalize the constraint loss by the length of the sequence so
that it does not grow unbounded with sequence size.  We recommend this
normalization as we found that it generally improves performance.
\subsection{Semantic Role labeling}
The $\closs(\outs,\lang^x)$ for SRL
factorizes into per-span constraints $\closs_i$.
For the $i$th span $s_i$, if $s_i$ is consistent with any node in the parse tree, 
$\closs_i(s_i,\lang^x)=0$, otherwise $\closs_i(s_i,\lang^x)=1/n_{s_i}$ where $n_{s_i}$ is defined as the number of tokens in $s_i$.
Overall,

\begin{equation*}
\energy(\inps,\hat{\outs},\weights_\lambda)\closs(\hat{\outs},\lang^x)=\sum_{i=1}^k \closs(s_i,\lang^x)\energy(\inps,s_i,W_\lambda)
\end{equation*}
where $k$ is number of spans on output $\hat{\outs}$.

More precisely, for a span $s$ to be ``consistent with a parse node'' we mean the
following.  Let $t_i\in T$ be a node in the parse tree $T$ and let $s^t_i$ be the
span of text implied by the descendents of the node $t_i$.  Let $S^T=\{s^t_i\}$
be the set of spans implied by all nodes in the parse tree $T$.  We say that a span
of text $s$ is consistent with the parse tree $T$ if and only if $s \in S^T$.

\subsection{Syntactic Parsing}
\label{appdx:parse-constraint}
Let $m_\inps$, $n$ be the number of input and output tokens, respectively, $\ncount_{i=1}^n(b(i))$ be the function that counts the number of
times proposition $b(i)$ is true for $i=1,\dots,n$. Now, define the following loss
\begin{align*}
\closs(\outs,\lang^x) =\frac{1}{m_\inps + n}\left\{\left|m_\inps - \ncount_{i=0}^n(y_i=s)\right|
+ \sum_{i}^n\max\left(0,\ncount_{j=0}^i(y_j=r) - \ncount_{j=0}^i(y_j\in\{s,!\})\right)  \right\}.
\end{align*}

The first term provides loss when the number or shifts equals the number of
input tokens, the second term provides loss when attempting to reduce
an empty stack and the third term provides loss when the number of
reduces is not sufficient to attach every lexical item to the tree.

\subsection{Transduction}

For the transducer we chose for our experiment,
$\lang_S$ is \texttt{(az|bz)}$^\star$ and the target language
$\lang_T$ is \texttt{(aaa|zb)}$^\star$. The transducer is defined to
map occurrences of \texttt{az} in the source string to \texttt{aaa} in
the target string, and occurrences of \texttt{bz} in the source string
to \texttt{zb} in the target string. 

For the provided transduction function,
the number of \texttt{a}'s in the output should be three times the
number of \texttt{a}'s in the input.
To express this constraint, we define following
constraint loss $\closs$, a length-normalized quadratic
\begin{align*}
\closs(\outs,\lang^x)  = (3x_a-y_a)^2/(m+n) 
\end{align*}
that is zero when the number of \texttt{a}'s in
the output ($y_a$) is exactly three times the number in the input
($x_a$) with $m$,$n$ denoting input length, output length, respectively.

\section{Analyzing the behavior of different inference procedures in
	the presence of noisy constraints}
\label{apdix:noisySRL}
\begin{table*}[h]
	\small
	\centering
	\begin{tabular}{|c|r|r|r|r|}
		\hline
		\multirow{1}{*}{Decoding} & \multirow{1}{*}{Precision} & \multirow{1}{*}{Recall} & \multirow{1}{*}{F1-score} & \multirow{1}{*}{Exact Match (\%)} \\
		\hline
		Viterbi & 84.03 & 84.78 & 84.40 & 69.37\\
		\hline
		& \multicolumn{4}{c|}{Noisy constraints} \\ 
		\cline{2-5}
		\multirow{1}{*}{\Astar} & 78.13 (-5.90) & 76.85 (-7.93) & 77.48 (-6.92) & 58.30 (-11.70) \\
		\multirow{1}{*}{GBI} & 85.51 (+1.48) & 84.25 (-0.53) & \textbf{84.87} (\textbf{+0.47}) & 68.45 (-0.92)\\
		\hline
		& \multicolumn{4}{c|}{Noise-free constraints} \\
		\cline{2-5}
		\multirow{1}{*}{\Astar} & 84.19 (+0.16) & 84.83 (+0.05) & 84.51 (+0.11) & 70.52 (+1.15) \\
		\multirow{1}{*}{GBI} & 85.39 (+1.36) & 85.88 (+1.10) & \textbf{85.63} (\textbf{+1.23}) & 71.04 (+1.67)\\
		\hline  
	\end{tabular}
	\caption{Comparison of different inference procedures: Viterbi, \Astar \cite{he2017deep} and GBI with noisy and noise-free constraints. Note that the (+/-) F1 are reported w.r.t Viterbi decoding on the same column.}
	\label{tab:struct-predict}
\end{table*}

Table \ref{tab:struct-predict} reports the performance of GBI and
\Astar in the presence of noisy constraints. We can see that the
overall performance (F1-score) for \Astar drops drastically ($-6.92$)
in the presence of noisy constraints while we still see gains with GBI
($+0.47$). 
We further analyze the improvement of GBI
by looking at the precision and recall scores individually. 
We see that recall drops slightly for GBI which suggests that noisy constraints do
inhibit predicting actual argument spans. On the other hand, we see
that precision increases significantly. After analyzing predicted
argument spans, we noticed that GBI prefers to predict no argument
spans instead of incorrect spans in the presence of noisy constraints
which leads to an increase in precision. Thus, GBI provides
flexibility in terms of strictness with enforcing constraints which
makes it robust to noisy constraints. On the other hand, constrained-\Astar decoding algorithm 
is too strict when it comes to enforcing noisy constraints
resulting in a significant drop of both precision and recall.

\section{Analyzing the runtime of different inference procedures with varying dataset sizes and genres}
\label{apdix:runtimeSRL}


\begin{table*}[h]
	\centering
	\begin{tabular}{|c|c|r|c|r|r|r|}
		\hline
		\multirow{2}{*}{Network}& \multirow{2}{*}{Genre(s)} & \multirow{2}{*}{No. of examples} & Failure &  \multicolumn{3}{c|}{Inference time (approx. mins)} \\\cline{5-7}
		&  &  & rate (\%) & Viterbi & GBI & \Astar \\ \hline
		SRL-100 & All & 25.6k & 9.82 & 109 & 288 & 377\\ \hline
		\multirow{5}{*}{SRL-NW} & BC & 4.9k & 26.86 & 23 & 110 & 117\\ \cline{2-7}
		& BN & 3.9K & 18.51 &18 & 64 & 100\\ \cline{2-7}
		& PT & 2.8k & 10.01 & 8 & 19 & 15 \\ \cline{2-7}
		& TC & 2.2k & 19.01 & 5 & 23 & 20\\ \cline{2-7}
		& WB & 2.3k & 20.32& 12 & 49 & 69\\ \hline
	\end{tabular}
	\caption{Comparison of runtime for difference inference
		procedures in the noise-free constraint setting: Viterbi, \Astar \cite{he2017deep} and GBI. For SRL-100 refer Table \ref{tab:srl} and SRL-NW is a model trained on NW genre.}
	\label{tab:runtime} 
\end{table*}

Table \ref{tab:runtime} reports the runtime for different inference
procedures with varying dataset sizes.
In general, we observe that GBI tends to be faster than
\Astar, especially when the dataset is large enough.
One exception is the BC domain where GBI is just slightly faster than
A*. We hypothesize it might be due to the difficulty of the constraint
violations as its failure rate is higher than usual.
GBI will spend more time searching for the correct output (more iterations) if it is harder to find the solution.

Also note that we explicitly set the max epoch $M$ for GBI after which it will stop iterating to avoid pathological cases. In our SRL experiments, we have set the max epochs to be 10 (GBI-10). To study its scalability, we ran GBI with max epoch set to 30 (GBI-30). The runtime increase to 556 mins for GBI-30 as opposed to 288 mins of GBI-10.
However, GBI-30 improves significantly in all accuracy metrics
compared to GBI-10: overall F1 (+0.34), F1 on failure set (+3.4),
exact match (+4.35\%), and conversion rate (+11.24\%). As can be
seen from the demonstration, there is a clear tradeoff between
runtime and accuracy as controlled by the maximum number of epochs $M$. The user can control $M$ by the runtime constraint the system has: lower $M$ when the serving time is most important and larger $M$ when accuracy is more important than the serving time.



\end{document}